\documentclass[runningheads]{llncs}

\usepackage{eccvabbrv}
\usepackage[mobile]{eccv}

\usepackage{graphicx}
\usepackage{booktabs}
\usepackage[accsupp]{axessibility}  
\usepackage[pagebackref,breaklinks,colorlinks,citecolor=eccvblue]{hyperref}
\usepackage[dvipsnames]{xcolor}

\usepackage{url}
\usepackage{wrapfig}

\usepackage{epsfig}
\usepackage{amsmath}
\usepackage{amssymb}
\usepackage{xspace}
\usepackage{amstext}
\usepackage{algorithm}
\usepackage{algorithmic}
\usepackage{multirow}
\usepackage{setspace}

\usepackage{newfloat}
\usepackage{listings}
\usepackage{bibentry}
\usepackage{bbding}
\usepackage{bm}
\usepackage{bbm}
\definecolor{dark-red}{rgb}{0.4,0.15,0.15}
\definecolor{dark-blue}{rgb}{0.15,0.15,0.4}
\definecolor{medium-blue}{rgb}{0,0,0.5}
\definecolor{cvprblue}{rgb}{0.21,0.49,0.74}

\usepackage[marginal]{footmisc}

\usepackage{orcidlink}

\begin{document}

\title{Self-Cooperation Knowledge Distillation for Novel Class Discovery}

\titlerunning{}

\author{Yuzheng Wang\inst{1} \and
Zhaoyu Chen\inst{1} \and
Dingkang Yang\inst{1} \and
Yunquan Sun\inst{1},  \\ and
Lizhe Qi\inst{1, \dag}
}

\authorrunning{ }

\institute{Shanghai Engineering Research Center of AI \& Robotics, Academy for Engineering \& Technology, Fudan University \\
\email{yzwang20@fudan.edu.cn}}

\maketitle

\begin{abstract}

Novel Class Discovery (NCD) aims to discover unknown and novel classes in an unlabeled set by leveraging knowledge already learned about known classes.
Existing works focus on instance-level or class-level knowledge representation and build a shared representation space to achieve performance improvements.
However, a long-neglected issue is the potential imbalanced number of samples from known and novel classes, pushing the model towards dominant classes.
Therefore, these methods suffer from a challenging trade-off between reviewing known classes and discovering novel classes.
Based on this observation, we propose a Self-Cooperation Knowledge Distillation (SCKD) method to utilize each training sample (whether known or novel, labeled or unlabeled) for both review and discovery.
Specifically, the model’s feature representations of known and novel classes are used to construct two disjoint representation spaces.
Through spatial mutual information, we design a self-cooperation learning to encourage model learning from the two feature representation spaces from itself.
Extensive experiments on six datasets demonstrate that our method can achieve significant performance improvements, achieving state-of-the-art performance.
\footnote{\dag $\,$ Corresponding Author} 
\end{abstract}

\begin{figure*}[t]
	\centering
	\includegraphics[width=\textwidth]{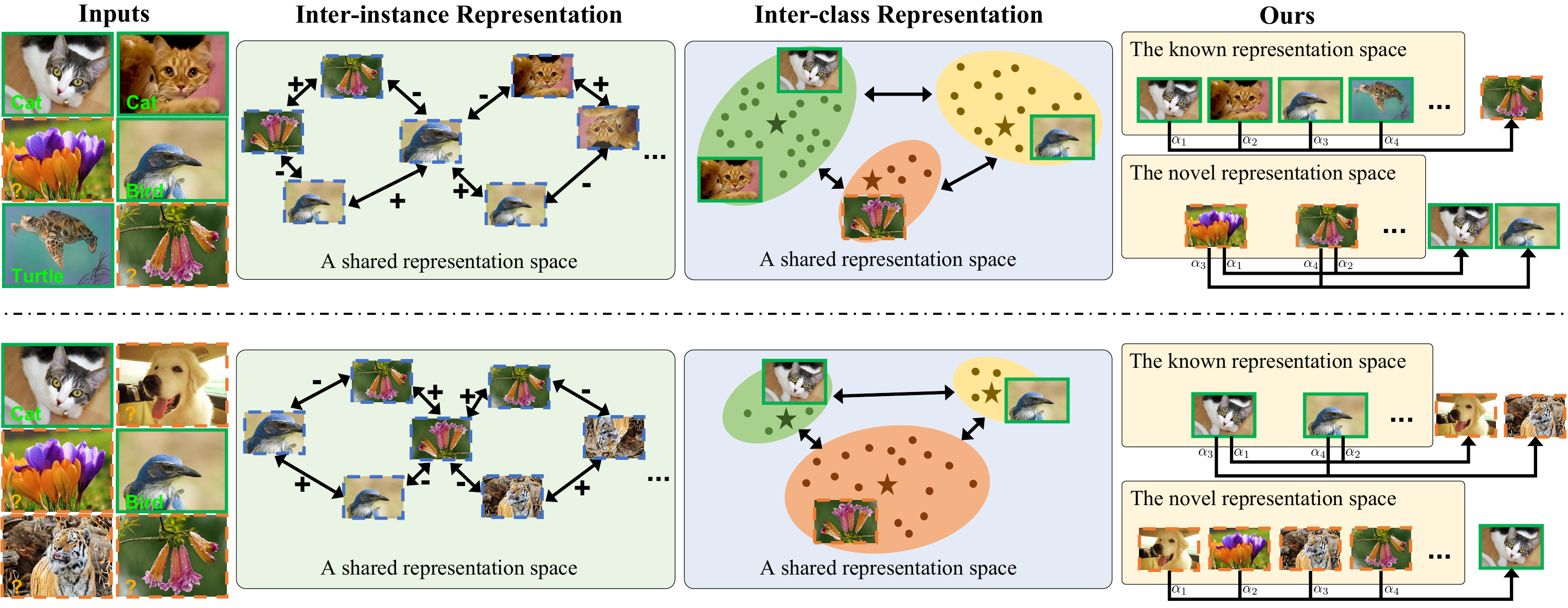}
	\caption{Diagram of more known samples than novel samples (upper) \& more novel samples than known samples (lower). 
    \textbf{Inter-instance methods} aim to explore relationships among instances via contrastive learning (KCL \cite{hsu2017learning}, MCL \cite{hsu2019multi}, NCL \cite{zhong2021neighborhood}, GCD \cite{vaze2022generalized}, DCCL \cite{pu2023dynamic}), rank statistics (RS \cite{han2020automatically}, DualRank \cite{zhao2021novel}), consistency and regularization (DTC \cite{han2019learning}, ComEx \cite{yang2022divide}, SimGCD \cite{wen2023parametric}), and example mixing (OpenMix \cite{zhong2021openmix}). \textbf{Inter-class methods} aim to explore relationships among multiple classes (IIC \cite{li2023modeling}, rKD \cite{peiyan2023class}). 
    \textbf{Our method} builds a self-cooperation pipeline for model learning.}
	\label{fig1}
    \vspace{-0.2cm}
\end{figure*}

\section{Introduction}
\label{sec:intro}

Deep learning, as a powerful and reliable tool, is increasingly popular in established and emerging fields of artificial intelligence \cite{he2016deep,redmon2018yolov3,yang2024robust,liu2023stochastic,liu2023learning,liu2023improving}.
However, a considerable part of this success comes from the high-quality labeled training data for each class \cite{yang2023context,ge2022zoom,yang2023aide,yang2024how2comm}, which is expensive \cite{chen2022towards,wang2023adversarial} or even impossible \cite{wang2023explicit,wang2024out}.
For lower cost, an essential subject is using unlabeled data and training scalable models \cite{laine2016temporal,he2020momentum,chen2020simclr,wang2024confounded,wang2023sampling}.
In this situation, Novel Class Discovery (NCD) task \cite{hsu2017learning, hsu2019multi} has been proposed.
NCD aims to train scalable models to discover unknown classes in unlabeled sets with the help of semantic representations learned in known classes in labeled sets (ideally the models should not forget known classes).
Note that this task is similar but different from Semi-Supervised Learning (SSL) \cite{tarvainen2017mean} and Zero-Shot Learning (ZSL) \cite{wang2019survey}.
The former assumes the labeled and unlabeled data share the same distribution and label space.
The latter relies on additional descriptions (usually derived from human annotation) of unknown classes.
With fewer dependencies and the advantages of open-world practicality, NCD has recently gained significant attention \cite{troisemaine2023novel}.

Existing NCD methods focus on establishing a shared representation space for known and novel instances or classes.
However, a long-neglected issue is the imbalanced number of samples from known and novel classes pushes the model toward the dominant party, making it challenging to trade-off between reviewing known classes and discovering novel classes.
As shown in \cref{fig1}, we give a diagram of the mainstream NCD methods.
Firstly, few novel samples and massive known samples lead to the pseudo-supervised information about the novel classes being overlooked due to a slight proportion.
Relatively, massive novel samples and few known samples lead to the information of the known classes being continuously marginalized and gradually approaching unsupervised learning.
In either case, it makes the learned feature representations biased toward one or the other.

\begin{wrapfigure}{r}{6.9cm}
	\centering
	\includegraphics[scale=0.19]{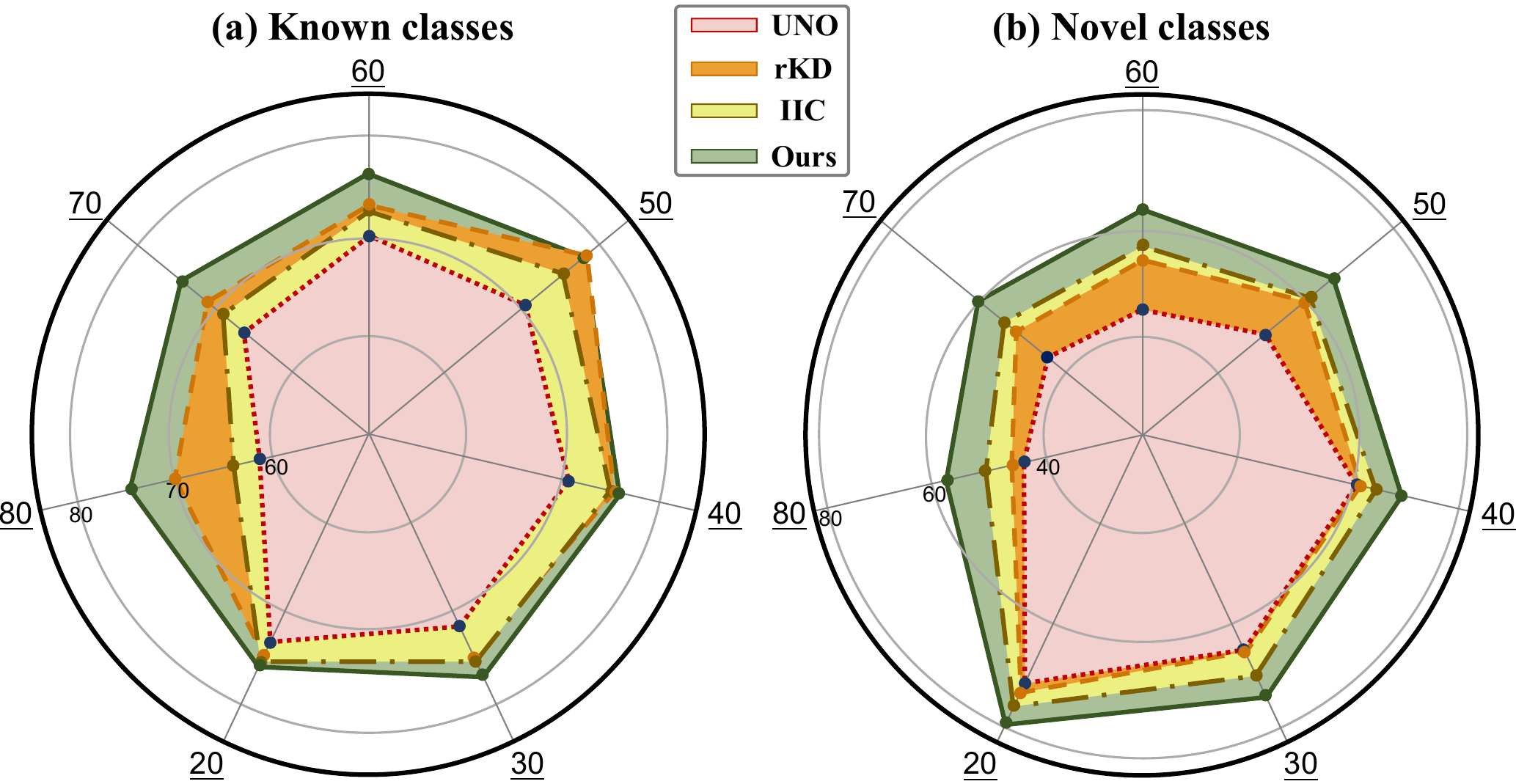}
	\caption{Performance overview on testing subsets of CIFAR100 with an increasing number of unlabeled classes. The \underline{underline} numbers represent the number of \textbf{novel classes}. 
    We report the accuracy of both (a) known and (b) novel classes.}
	\label{fig3}
	\vspace{-0.5cm}
\end{wrapfigure}

More intrigued, we evaluate the performance of the competitive NCD methods when dealing with different numbers of known and novel classes and samples on CIFAR100.
As shown in \cref{fig3}, as the number of novel classes increases, the prediction accuracy of the current methods for known classes drops significantly. 
The performance of novel classes directed by known classes is also inevitably affected.
Such results show that imbalanced samples affect existing methods.

Based on above observation, a natural idea emerged: 
\textit{Can we use all sample information to review known classes and discover novel classes simultaneously}?
As shown in \cref{fig1}, we establish and maintain non-overlapping representation
spaces for the known and novel classes respectively.
On this basis, we propose Self-Cooperation Knowledge Distillation (SCKD).
Specifically, we treat the model's feature about known and novel classes as pairwise mutual information and encourage the model to perform self-cooperation learning between the two feature representation spaces.
For instance, when few known samples are not enough to support knowledge review, the model can use a large number of unlabeled samples of novel classes to help review known classes.
Based on these, the model can cooperatively discover novel classes and review known classes.
Specifically, the primary contributions are summarized as follows:

\begin{itemize}
    \item We consider a practical but long-neglected challenge in the NCD task, \ie, the imbalanced number of samples from known and novel classes, making it difficult to balance reviewing known classes and discovering novel classes.
    \item We propose a simple yet effective SCKD method. SCKD can associate every sample for simultaneously reviewing known classes and discovering novel classes by building a cooperative learning paradigm.
    \item Extensive experiments on six benchmark datasets for novel class discovery show that the proposed method performs competitively and outperforms the state-of-the-art methods, demonstrating the effectiveness of SCKD.
\end{itemize}

\section{Related Work}

\noindent \textbf{Novel Class Discovery.}
The Novel Class Discovery (NCD) task has two main goals: 1) maintaining the model's recognition performance for known classes and 2) promoting the model's discovery of class-disjoint novel classes.
The early methods predate the definition of NCD but focus on similar real-world needs.
KCL \cite{hsu2017learning} and MCL \cite{hsu2019multi} use pairwise semantic similarity learning to distinguish whether instances belong to the same class and reconstruct semantic clusters to achieve knowledge transfer.
Han \etal \cite{han2019learning} standardize the description of the NCD task for the first time and propose the DTC method to observe and learn the temporal ensembling and consistency at different training stages to transfer knowledge from known to novel classes.
Subsequent methods also focus on maintaining the same shared representation for known and novel samples and classes, focusing on transferring known knowledge to novel classes.
Specifically, RS \cite{han2020automatically} learns on labeled and unlabeled samples through self-supervision learning and then proposes ranking statistics to measure the similarity of two data in the representation space.
DualRank \cite{zhao2021novel} expands RS and proposes the dual ranking statistics to ensure knowledge transfer and consistency learning at global and local levels.
NCL \cite{zhong2021neighborhood} introduces the explicit contrastive learning training objectives to learn the consistency between instances and their related views and neighbors in a shared representation space.
OpenMix \cite{zhong2021openmix} refers to the Mixup \cite{zhang2017mixup} to mix labeled and unlabeled samples to fuse and distinguish known and novel classes in the representation space.
Joint \cite{jia2021joint} uses category discrimination to augment contrastive learning and employs the Winner-Take-All (WTA) hashing algorithm on the shared representation space for cluster assignments.
UNO \cite{fini2021unified} proposes a unified training objective to use labeled samples for known category review and unlabeled samples for novel category discovery.
IIC \cite{li2023modeling} proposes to learn inter-class and intra-class constraints to establish instance-level and inter-class associations in the shared representation space to promote the distinction of novel instances and classes.
rKD \cite{peiyan2023class} introduces class relationship representation, transferring knowledge based on the predicted distribution of the pre-trained model on known classes to assist in learning novel classes.
In addition, some related tasks are proposed.
ComEx \cite{yang2022divide} focuses on generalized NCD and proposes using multiple regularizations for known and novel classes in the representation space.
GCD \cite{vaze2022generalized} defines Generalized Category Discovery to relax the settings of NCD and proposes a contrastive learning and category number estimation method.
On this basis, DCCL \cite{pu2023dynamic} and SimGCD \cite{wen2023parametric} respectively optimize the contrastive learning paradigm and pseudo-label synthesis method through entropy regularization in GCD \cite{vaze2022generalized}.

To our best knowledge, the existing methods do not fully consider the potential imbalanced number of samples from known and novel classes, leading the model toward dominant classes and ignore others.
Looking at NCD from the perspective of long-term and practical applications (\eg, recommendation system \& scalable foundation model), models may be encouraged to continuously learn novel classes of knowledge and inevitably lead to the eventual dilution and marginalization of known knowledge with the popularity of a shared representation space.
Instead, we maintain distinct representation spaces and propose cooperative learning for known and novel classes to alleviate this issue.

\noindent \textbf{Self Knowledge Distillation.}
Self Knowledge Distillation (Self-Distillation) uses the knowledge of the target model itself to guide the optimization of the model, which can be regarded as a special case of Knowledge Distillation \cite{allen2020towards}.
BAN \cite{furlanello2018born} models the temporal sequence in the model optimization process and uses the knowledge of the model at the previous moment to guide learning at the current moment.
SAD \cite{hou2019learning} allows the model to utilize its attention maps as distillation targets for its lower layers and obtain substantial performance improvements without additional supervision or labels.
Zhang \etal \cite{zhang2019your} layer the network into sections and add multiple bottleneck layers to realize the guidance of the deeper layers to the lower layers.
FRSKD \cite{ji2021refine} proposes a feature refinement self knowledge distillation method and guides the establishment of feature map distillation at the same level through integrating and refining each deep feature layer.
DLB \cite{shen2022self} proposes a training strategy to rearrange sequential sampling by limiting half of each mini-batch to coincide with the previous iteration. Meanwhile, the remaining half coincides with upcoming iterations.

Different from previous self knowledge distillation methods, we propose a customized self knowledge distillation technology for the NCD task, which aims to use the knowledge of all training samples of a mini-batch for both known class review and novel class discovery.
Our distillation process attempts to link the training objectives from two subsets for cooperative learning rather than maintaining separate training objectives like previous methods.
Notably, our method requires no additional storage overhead to retain training information for multiple training stages (epochs).

\begin{figure*}[t]
	\centering
	\includegraphics[width=\textwidth]{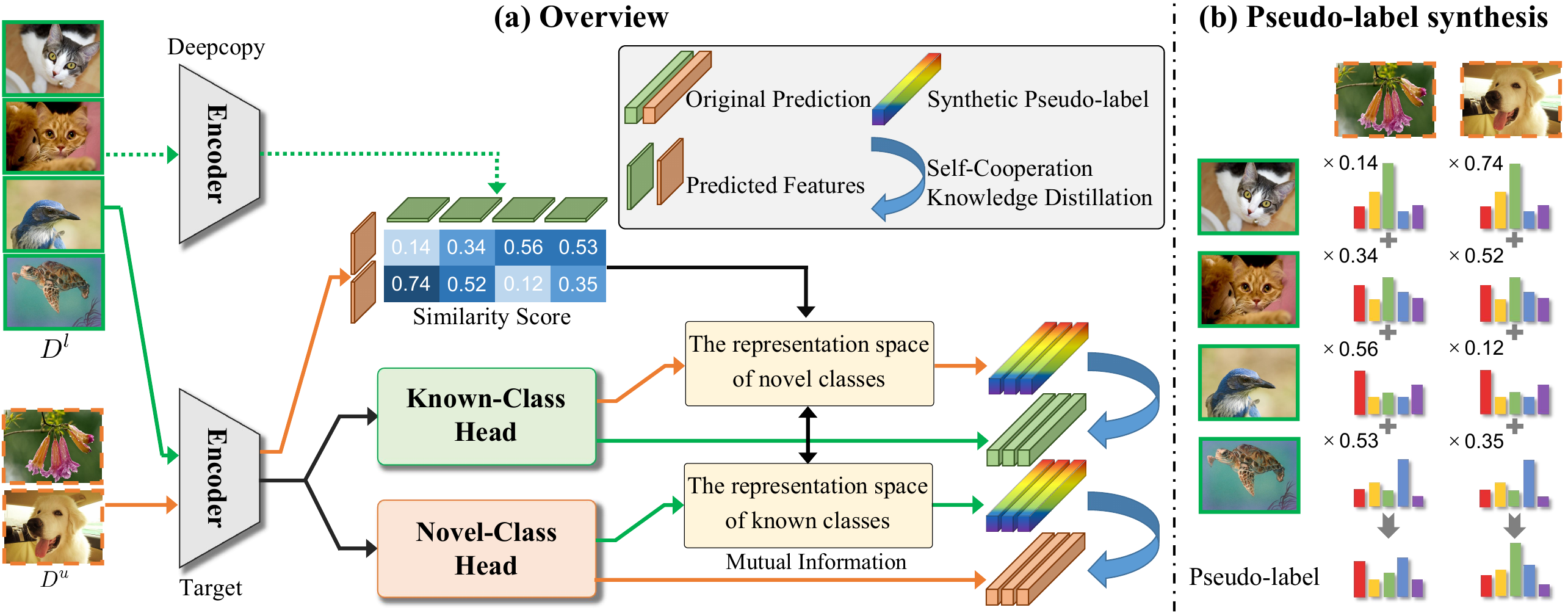}
	\caption{The overview of (a) our proposed Self-Cooperation Knowledge Distillation and (b) a schematic diagram of pseudo-label synthesis. 
    In (a), we define arrows with different colors and line shapes to distinguish data flows. 
    First, a similarity score matrix is constructed for the spatial mutual information between labeled and unlabeled samples. 
    Then, through the pseudo-label synthesis modules, known samples provide information for discovering novel classes and novel class samples provide information for reviewing known classes.
    In (b), the pseudo-supervised information comes from the logits prediction and weight from similarity scores. 
    The self-distillation process combines discovering novel classes and reviewing known classes to overcome the potential imbalanced number of samples from known and novel classes.}
	\label{fig2}
\end{figure*}

\section{Methodology}

\subsection{Preliminaries}

In the NCD task, a mini-batch training set contains a labeled subset of known classes $D^{l} =\left \{ (\bm{x}^{l}_1,\bm{y}^{l}_1),\dots,(\bm{x}^{l}_N,\bm{y}^{l}_N)  \right \} $ and an unlabeled subset of novel classes $D^{u} =\left \{ \bm{x}^{u}_1,\dots,\bm{x}^{u}_M \right \} $.
Each $\bm{x}^{l}_i$ in $D^{l}$ or $\bm{x}^{u}_i $ in $D^{u}$ is an input image.
And $\bm{y}^l_i\in \mathcal{Y}^l =\left \{ 1,\dots ,C^l \right \} $ is the real categorical label of $\bm{x}^{l}_i$.
Following the mainstream setting \cite{fini2021unified,li2023modeling}, the set of $C^l$ labeled classes is assumed to be disjoint from the set of $C^u$ unlabeled classes, and the number of latent unlabeled classes $C^u$ is assumed to be known as a priori.
The core purpose of NCD is to predict test instances as labeled classes or unlabeled clusters to learn matching composite label sets $ \mathcal{Y} =\left \{ 1,\dots,C^l,C^l+1,\dots,C^l+C^u  \right \} $.
Notably, there may be significant differences between $C^l$ and $C^u$, making it difficult to review known classes and discover novel classes simultaneously.
Based on the above setting, various methods build semantic knowledge and leverage the knowledge about labeled known classes to discover entangled latent classes.

Under the above settings, we propose a Self-Cooperation Knowledge Distillation (SCKD) method to associate the discovery of novel classes with the review of known classes (see \cref{fig2}) to alleviate the adverse effects of imbalanced number of samples from known and novel classes.
In the following sections, we first introduce the SCKD process and the distillation training objective in Sec.~\ref{sec3.2}.
Then, we summarize the overall training objective in Sec.~\ref{sec3.3}.

\subsection{Self-Cooperation Knowledge Distillation}
\label{sec3.2}


Our target model consists of a widely accepted architecture for NCD, \ie, an encoder $E$ and two classification heads $h^l$ and $h^u$.
The encoder $E$ converts the input image into encoded feature vectors.
The known-class head $h^l$ is implemented as a linear classifier with $C^l$ output dimensions.
The novel-class head $h^u$ is implemented as a multilayer perceptron (MLP) and a linear classifier with $C^u$ output dimensions.
Following existing methods \cite{fini2021unified,peiyan2023class}, we first conduct the standard supervised training on labeled known classes.
In addition, we deepcopy the trained encoder $E$ named replica encoder $E^r$ to ensure pure and unbiased feature information of the known classes.
Thereafter, the encoder $E$ remains trainable while the replica encoder $E^r$ remains frozen.
The total logits prediction for the input sample $\bm{x}_i$ comes from the concat representation predicted by the two classification heads as $\bm{l}_i = \left [ h^l(E(\bm{x}_i)), h^u(E(\bm{x}_i))   \right ], \bm{l}_i \in \mathbb{R}^{ C^l+C^u}  $.
The probability distribution can be obtained as $\bm{p}_i=\sigma(\bm{l}_i / \tau) $, where $\sigma$ is a softmax layer and $\tau$ is the softmax temperature.

For our SCKD method, based on the disjoint characteristic, we first divide the labeled and unlabeled samples into two groups to maintain separate representation spaces for known and novel classes respectively.
Then, we build information bridges from one sample to all samples in another group through a similarity score matrix.
Based on the score matrix, we obtain the soft pseudo labels about unlabeled samples from the known representation space and simultaneously obtain pseudo labels about labeled samples from the novel representation space.
Finally, we design a self knowledge distillation process to achieve cooperative learning of novel class discovery and known class review for unbiased representation learning.

\noindent \textbf{Similarity Score Matrix.}
To ensure the information balance between known and novel classes, we establish associations between labeled and unlabeled samples with a similarity score matrix so that all samples can contribute to discovering novel classes and reviewing known classes.
The similarity score measures the semantic correlation between labeled and unlabeled samples.
First, all labeled samples are fed to replica encoder $E^r$ to get the set of predicted features as $ \bm{v}^l = E^r(\bm{x}^{l}), \bm{v}^l \in \mathbb{R}^{N \times k} $, where $k$ is the output dimension of the encoder.
All unlabeled samples are fed to the target encoder $E$ to get the predicted features as $ \bm{v}^u = E(\bm{x}^{u}), \bm{v}^u \in \mathbb{R}^{M \times k} $.
The similarity matrix $S$ is calculated via the cosine similarity metric as follows:
\begin{equation}\label{eq1}
S_{ij}=  \cos \; (\bm{v}^l_i, \; \bm{v}^u_j), \;i\!=\!1,\dots,\!N \;\; \text{and} \;\; j\!=\!1,\dots,\! M.
\end{equation}
Then, the score matrix is obtained by the maximum value for normalization as follows:
\begin{equation}\label{eq2}
S = \text{Norm}(S) = \frac{S}{   \left |   max\left \{ S \right \}     \right | } , S \in \mathbb{R}^{N \times M}.
\end{equation}
It is worth noting that even if there is a serious imbalanced number of samples from the known and novel classes, marginal classes can still learn effectively from samples of dominant classes.

\noindent \textbf{Pseudo-Label Synthesis.}
After constructing two disjoint representation space and calculating the similarity matrix, we aim to synthesize pseudo-labels to form self-supervised information and encourage the model to utilize the spatial mutual information to overcome the imbalanced number of samples from known and novel classes.
All labeled and unlabeled samples are fed to the target model.
We collect the labeled and unlabeled predictions at each classification head.
For the novel-class head $h^u$, we extract information about labeled samples from the known representation space and collect the predicted logits $ \bm{l}^l_{uh} = h^u(E(x^l)), \;\bm{l}^l_{uh} \in \mathbb{R}^{N \times C^u}$.
Combined with the similarity score matrix $S$, we map the information of the known representation space into unlabeled pseudo-label representation, \ie, the process of novel class discovery using sample knowledge of known classes (see \cref{fig2}\textcolor{red}{b}).
The formalized mathematical expression can be summarized as follows:
\begin{equation}\label{eq3}
\hat{\bm{l}}^u_{uh} = \alpha \cdot S^T \!\cdot \bm{l}^l_{uh}, \;\; \hat{\bm{l}}^u_{uh} \in \mathbb{R}^{M \times C^u},
\end{equation}
where $\alpha$ is a label smoothing coefficient used to adjust the degree of mapping intervention between known and novel classes.
Similarly, for the known-class head $h^l$, we extract information about unlabeled samples from the novel representation space and collect the predicted logits $ \bm{l}^u_{kh} = h^l(E(x^u)), \; \bm{l}^u_{kh}  \in \mathbb{R}^{M \times C^l}$.
The information of the novel-class representation space is mapped into labeled pseudo-label representation as follows:
\begin{equation}\label{eq4}
\hat{\bm{l}}^l_{kh} = \alpha \cdot S \cdot \bm{l}^u_{kh}, \;\; \hat{\bm{l}}^l_{kh} \in \mathbb{R}^{N \times C^l}.
\end{equation}

\noindent \textbf{Self Knowledge Distillation Objectives.}
After obtaining the pseudo-label representation, we encourage models to learn consistent knowledge representations that connect known and novel representation spaces.
To utilize the information of known class samples in the novel-class discovery process, we collect the original predictions $\bm{l}^u_{uh} = h^u(E(x^u)), \;\bm{l}^u_{uh} \in \mathbb{R}^{M \times C^u} $ of the novel-class head $h^u $ for unlabeled samples.
The pseudo-label representation from the spatial mutual information is expected to guide updating the original prediction, \ie, a self-distillation process between labeled and unlabeled samples.
The self-distillation loss for consistency constraints can be computed as follows:
\begin{equation}\label{eq5}
\mathcal{L}_{k\to n}=\frac{1}{M} \sum_{1}^{M} KL(\bm{l}^u_{uh}  ,\hat{\bm{l}}^u_{uh} ),
\end{equation}
where $KL$ denotes the Kullback-Leibler divergence loss.
With the help of $\mathcal{L}_{k\to n}$, the target model can fully use all labeled and unlabeled samples to discover novel classes. 
Similarly, to utilize the information of novel class samples in the known-class review process, we constrain the original predictions $\bm{l}^l_{kh} = h^l(E(x^l)), \;\bm{l}^l_{kh} \in \mathbb{R}^{N \times C^l} $ of the known-class head $h^l $ for labeled samples and pseudo-labels from the unlabeled sample $\hat{\bm{l}}^l_{kh}$ as follows:
\begin{equation}\label{eq6}
\mathcal{L}_{n\to k}=\frac{1}{N} \sum_{1}^{N} KL(\bm{l}^l_{kh} ,\hat{\bm{l}}^l_{kh} ).
\end{equation}
$\mathcal{L}_{n\to k}$ encourages the target model to integrate all labeled and unlabeled samples for reviewing known classes.

Notably, previous methods focus on using labeled samples to review known classes and unlabeled samples to discover novel classes.
In contrast, our method builds a cooperative process to comprehensively use labeled samples (known classes) to guide the learning of unlabeled samples (discover novel classes). Simultaneously, it can use unlabeled samples (novel classes) to guide the learning of labeled samples (review known classes).
Therefore, our self-distillation method can cope with the challenge of the imbalanced number of samples from known and novel classes.
In addition, our method encourages the model to joint optimization of the known and novel classes, thus helping to avoid catastrophic forgetting of known classes and over-conservatism in novel class learning.
Finally, we combine the two parts of the training objectives to form our overall Self-Cooperation Knowledge Distillation loss as follows:
\begin{equation}\label{eq7}
\mathcal{L}_{SCKD} = \mathcal{L}_{k\to n} + \mathcal{L}_{n\to k}.
\end{equation}

\subsection{Overall Objective}
\label{sec3.3}

In addition to the proposed distillation loss, we also use the cross-entropy loss as follows:
\begin{equation}\label{eq8}
\mathcal{L}_{CE} = -\frac{1}{N+M}\sum_{1}^{N+M} \bm{y}_i \;\text{log} \; \bm{p}_i , \; \bm{y}_i\in \mathcal{Y},
\end{equation}
where $\bm{y}_i$ denotes label of image $\bm{x}_i$ and $\bm{p}_i$ denotes the probability distribution of the model for $\bm{x}_i$.
Following existing baselines \cite{fini2021unified}, $\bm{y}_i$ is one-hot-like ground truth for the labeled image.
For the unlabeled image, $\bm{y}_i$ is the model's pseudo label after Sinkhorn-Knopp regularization \cite{cuturi2013sinkhorn}, following mainstream NCD methods \cite{fini2021unified,peiyan2023class}.
Ultimately, our overall training objective is
\begin{equation}\label{eq9}
\mathcal{L} = \mathcal{L}_{CE} + \beta \cdot \mathcal{L}_{SCKD},
\end{equation}
where $ \beta$ is the loss trade-off parameter.

\section{Experiments}

\subsection{Experimental Setup}

\noindent \textbf{Datasets.} 
We evaluate the proposed method under NCD settings on both the generic image recognition datasets (including CIFAR10 \cite{krizhevsky2009learning}, CIFAR100 \cite{krizhevsky2009learning} and ImageNet-100 \cite{deng2009imagenet}) and the fine-grained datasets (including Stanford Cars \cite{krause20133d}, CUB \cite{wah2011caltech} and FGVC-Aircraft \cite{maji2013fine}).
Following UNO \cite{fini2021unified}, we divide CIFAR100 into two categories: 80/20 known/novel classes and 50/50 known/novel classes.
ImageNet-100 denotes randomly sub-sampling 100 classes from the ImageNet dataset.
The details of the dataset splits are shown in \cref{tab1}.

\noindent \textbf{Evaluation Metrics.} 
Following the mainstream NCD evaluation paradigm \cite{fini2021unified,li2023modeling}, we conduct our experiments using both \textbf{task-aware} and \textbf{task-agnostic} evaluation protocols.
For task-aware evaluation protocol, it is a priori information that each test sample comes from known or novel classes.
For task-agnostic evaluation protocol, above information about classes is unknown.
Following the widely recognized project, we use the accuracy measure for labeled samples and the average clustering accuracy for unlabeled samples.
The average clustering accuracy is defined as:
\begin{equation}\label{eq10}
\text{ClusterAcc}=\underset{perm\in P}{\max} \frac{1}{N}\sum_{i=1}^{N}  \mathbbm{1} \left \{ y_i= perm (\hat{y}_i)  \right \},
\end{equation}
where $y_i $ is the ground-truth label and $\hat{y}_i$ is predicted clusters of a test sample $\bm{x}^u_i \in D^u$.
$P$ is the set of all permutations computed with the Hungarian algorithm \cite{Hungarian}.
We report the results over 5 runs for ImageNet-100 and 3 runs for other datasets in subsequent experimental evaluations.

\begin{table*}[t]
\centering
\caption{The details of dataset splits involved in the experiments.}
\renewcommand\arraystretch{0.95}
\setlength{\tabcolsep}{1mm}
\scalebox{0.7}{
\begin{tabular}{@{}ccccccccc@{}}
\toprule
\multicolumn{2}{c}{Datasets} & CIFAR10 & CIFAR100-20 & CIFAR100-50 & ImageNet-100 & Stanford Cars & CUB & Aircraft \\ \midrule
\multirow{2}{*}{Known} &
  Images &
  25K &
  40K &
  25K &
  $ \approx$63.7K &
  $\approx$4.0K &
  $\approx$3.0K &
  $ \approx$3.3K \\
          & Classes          & 5       & 80          & 50          & 50           & 98            & 100 & 50            \\ \midrule
\multirow{2}{*}{Novel} &
  Images &
  25K &
  10K &
  25K &
  $ \approx$63.4K &
  $ \approx$4.1K &
  $ \approx$3.0K &
  $\approx$3.3K \\
          & Classes          & 5       & 20          & 50          & 50           & 98            & 100 & 50            \\ \bottomrule
\end{tabular}
}
\label{tab1}
\end{table*}

\begin{table*}[t]
\centering
\caption{Comparison with state-of-the-art methods on the unlabeled training subset, using task-aware evaluation protocol. 
\textbf{Bold} and \underline{underline} numbers denote the best and the second best results, respectively.}
\renewcommand\arraystretch{1}
\setlength{\tabcolsep}{1mm}
\scalebox{0.72}{
\begin{tabular}{@{}cccccccc@{}}
\toprule
Method     & CIFAR10  & CIFAR100-20 & CIFAR100-50 & ImageNet-100 & Stanford Cars & CUB      & Aircraft \\ \midrule
$k$-means \cite{macqueen1967some}     & 72.5$\pm$0.0 & 56.3$\pm$1.7    & 28.3$\pm$0.7    & 67.21        & 13.1$\pm$1.0      & 42.2$\pm$0.5 & 18.5$\pm$0.3      \\
KCL \cite{hsu2017learning}        & 72.3$\pm$0.2 & 42.1$\pm$1.8    & -           & -            & -             & -        & -             \\
MCL \cite{hsu2019multi}        & 70.9$\pm$0.1 & 21.5$\pm$2.3    & -           & -            & -             & -        & -             \\
DTC \cite{han2019learning}       & 88.7$\pm$0.3 & 67.3$\pm$1.2    & 35.9$\pm$1.0    & -            & -             & -        & -             \\
RS+ \cite{han2020automatically} & 91.7$\pm$0.9 & 75.2$\pm$4.2    & 44.1$\pm$3.7    & -            & 36.5$\pm$0.6      & 55.3$\pm$0.8 & 38.4$\pm$0.6      \\
OpenMix \cite{zhong2021openmix}   & 95.3     & 87.2        & -           & 74.76        & -             & -        & -             \\
NCL \cite{zhong2021neighborhood}       & 93.4$\pm$0.5 & 86.6$\pm$0.4    & 52.7$\pm$1.2    & -            & 43.5$\pm$1.2      & 48.1$\pm$0.9 & 43.0$\pm$0.5      \\
Joint  \cite{jia2021joint}    & 93.4$\pm$0.6 & 76.4$\pm$2.8    & -           & -            & -             & -        & -             \\
DualRank \cite{zhao2021novel}   & 91.6$\pm$0.6 & 75.3$\pm$2.3    & -           & -            & -             & -        & -             \\
ComEx \cite{yang2022divide}     & 93.6$\pm$0.3 & 85.7$\pm$0.7    & 53.4$\pm$1.3    & -            & -             & -        & -             \\
UNO  \cite{fini2021unified}      & 93.3$\pm$0.4 & 90.5$\pm$0.7    & 62.3$\pm$1.4    & 79.56        & 49.8$\pm$1.4      & 59.2$\pm$0.4 & 52.1$\pm$0.7      \\
GCD   \cite{vaze2022generalized}     & -        & -           & -           & 71.75        & 42.6$\pm$0.4      & 56.4$\pm$0.3 & 49.5$\pm$1.0      \\
SimGCD  \cite{wen2023parametric}   & -        & -           & -           & \underline{81.92}        & 50.2$\pm$0.5      & 62.3$\pm$0.4 & 53.6$\pm$1.1      \\
IIC  \cite{li2023modeling}      & \textbf{99.1$\pm$0.0} & \underline{92.4$\pm$0.2}    & \underline{65.8$\pm$0.9}    & 80.24        & \underline{55.2$\pm$0.7}      & \underline{71.3$\pm$0.6} & \underline{56.0$\pm$0.8}      \\
rKD   \cite{peiyan2023class}     & 93.5$\pm$0.3 & 91.2$\pm$0.1    & 65.3$\pm$0.6    & 80.94        & 53.5$\pm$0.8      & 65.7$\pm$0.6 & 55.8$\pm$0.9      \\ \midrule
\textbf{SCKD}       & \underline{95.6$\pm$0.2} & \textbf{92.6$\pm$0.6}    & \textbf{68.2$\pm$0.4}    & \textbf{82.18}        & \textbf{56.8$\pm$0.7}      & \textbf{73.1$\pm$0.4}  & \textbf{56.5$\pm$0.7}      \\ \bottomrule
\end{tabular}
}
\label{tab2}
\end{table*}

\noindent \textbf{Implementation Details.} 
Following the mainstream NCD method \cite{han2020automatically,fini2021unified}, by default, we adopt a two-stage training strategy, \ie, first a supervised training stage, and then a discovery training stage.
The replica encoder $E^r$ is derived from the pre-trained model after the first stage and frozen in the second stage.
For CIFAR10, CIFAR100 and ImageNet-100, we use ResNet18 \cite{he2016deep} as our backbone network.
We first pre-train on labeled samples for 100 epochs and then train on all labeled and unlabeled samples for 500 epochs, following UNOv2 \cite{fini2021unified}.
The learning rate initially increases from 0.001 to 0.4 during the first 10 epochs and then decreases to 0.001 at 500 epochs with a cosine annealing schedule. 
We adopt the SGD optimizer with the momentum as 0.9, weight decay as $ 1.5\times10 ^{-4}$.
For a fair comparison, we introduce the multi-head technique same with UNO \cite{fini2021unified} and employ four heads.
For Stanford Cars, CUB, FGVC-Aircraft datasets, we utilize DINO pre-trained ViT-B-16 \cite{caron2021emerging} as the backbone, use the output of \texttt{[CLS]} token with a dimension of 768 as the encoder, and only finetune its last block.
Besides, we use the AdamW optimizer and employ two heads.
The learning rate initially increases from 0.0001 to 0.001 during ten epochs and then decreases to 0.0001 at 100 epochs with a cosine annealing schedule.
We first pre-train on labeled samples for 50 epochs and then train on all labeled and unlabeled samples for 100 epochs, following UNO \cite{fini2021unified}, IIC \cite{li2023modeling}, and rKD \cite{peiyan2023class}.
For all datasets, the softmax temperature $\tau$ is set to 0.1, the temperature of knowledge distillation is 1, the hyperparameters $\alpha$ is 0.1, and $\beta$ is 0.5.
Besides, we set the batch size to 512 and apply the multi-crop strategy (\eg, random crop, flip, color jittering, and grey-scale), following UNO \cite{fini2021unified}, IIC \cite{li2023modeling} and rKD \cite{peiyan2023class}.
For the Sinkhorn-Knopp algorithm, we inherit the hyperparameters $n\_{iter}=3$ and $\epsilon =0.05$.

\subsection{Performance Comparison}
\label{sec4.2}

We compare our method with state-of-the-art NCD models, including: KCL \cite{hsu2017learning}, MCL \cite{hsu2019multi}, DTC \cite{han2019learning}, RS+ \cite{han2020automatically}, OpenMix \cite{zhong2021openmix}, NCL \cite{zhong2021neighborhood}, Joint \cite{jia2021joint}, DualRank \cite{zhao2021novel}, UNO \cite{fini2021unified}, IIC \cite{li2023modeling}, and rKD \cite{peiyan2023class}.
Furthermore, we reproduce the training objectives in GCD methods under NCD settings, including ComEx \cite{yang2022divide}, GCD \cite{vaze2022generalized}, and SimGCD \cite{wen2023parametric}.

\noindent \textbf{Training Subset.}
We first report the average clustering accuracy on the training split of the unlabeled subset, using the task-aware evaluation protocol (a common practice in the literature \cite{han2019learning,fini2021unified}).
As shown in \cref{tab2}, our SCKD method outperforms the other SOTA methods on most benchmarks, except CIFAR10.
We suspect that a possible reason is that the data distribution in CIFAR10 is relatively simple, and the semantic information is not rich enough, which may reduce the effectiveness of the semantic similarity-based module in SCKD.
In addition, for other semantically rich benchmark datasets, our method can achieve stable gains (higher average performance and smaller standard deviation).

\noindent \textbf{Testing Subset.}
We also compare the performance of the testing subset with labeled and unlabeled samples, using the task-agnostic evaluation protocol.
The results are shown in \cref{tab3}.
Our method achieves the best results on almost all benchmarks for both known and novel classes.
A well-founded finding is that our method significantly improves the model's performance for novel classes.
In NCD tasks, the model is often more difficult to learn novel classes (from the test results). Previous methods often only used unlabeled samples for novel class learning, losing the information of labeled samples. 
In contrast, our SCKD method additionally incorporates information from labeled samples without introducing additional forward overhead, and the proposed cooperative learning paradigm also helps both known and novel class learning.

\begin{table*}[t]
\centering
\caption{Comparison with state-of-the-art methods on the testing subset, using task-agnostic evaluation protocol. }
\renewcommand\arraystretch{1}
\setlength{\tabcolsep}{0.85mm}
\scalebox{0.7}{
\begin{tabular}{@{}c|ccc|ccc|ccc|ccc|ccc@{}}
\toprule
\multirow{2}{*}{Method} &
  \multicolumn{3}{c|}{CIFAR100-50} &
  \multicolumn{3}{c|}{CIFAR100-80} &
  \multicolumn{3}{c|}{Stanford Cars} &
  \multicolumn{3}{c|}{CUB} &
  \multicolumn{3}{c}{Aircraft} \\
           & Known & Novel & All  & Known & Novel & All  & Known & Novel & All  & Known & Novel & All  & Known & Novel & All  \\ \midrule
RS+ \cite{han2020automatically} & 69.7  & 40.9  & 55.3 & 71.2  & 56.8  & 68.3 & 81.8  & 31.7  & 56.3 & 80.7  & 51.8  & 66.1 & 66.4  & 36.5  & 51.5 \\
NCL \cite{zhong2021neighborhood}       & 72.4  & 25.7  & 49.0 & 72.7  & 41.6  & 66.5 & 83.5  & 24.4  & 53.4 & 79.8  & 13.1  & 46.3 & 62.8  & 26.5  & 44.6 \\
UNO \cite{fini2021unified}       & 71.5  & 50.7  & 61.1 & 73.2  & 73.1  & 73.2 & 81.7  & 46.7  & 63.9 & 78.7  & 62.1  & 70.3 & 71.2  & 52.4  & 61.8 \\
rKD  \cite{peiyan2023class}      & \textbf{78.6}  & 59.4  & \underline{69.0} & 75.2  & 76.4  & 75.4 & \underline{83.9}  & 51.3  & \underline{67.3} & \underline{81.1}  & 67.5  & 74.2 & \underline{72.2}  & 55.2  & \underline{63.7} \\
IIC \cite{li2023modeling}       & 75.1  & \underline{61.0}  & 68.1 & \underline{75.9}  & \underline{78.4}  & \underline{76.4} & 82.7  & \underline{51.4}  & 66.8 & 80.2  & \underline{69.4}  & \underline{74.8} & 71.2  & \underline{55.5}  & 63.3 \\
\textbf{SCKD}       & \underline{77.9}  & \textbf{62.6}  & \textbf{70.3} & \textbf{76.1}  & \textbf{79.7}  & \textbf{76.8} & \textbf{84.3}  & \textbf{52.0}  & \textbf{67.9} & \textbf{81.2}  & \textbf{72.3}  & \textbf{76.8} & \textbf{72.4}  & \textbf{56.1}  & \textbf{64.3} \\ \bottomrule
\end{tabular}
}
\label{tab3}
\end{table*}

\begin{table*}[t]
\centering
\caption{Experimental results with an increasing number of unlabeled classes on CIFAR100. Results are reported on the testing subset about both known and novel classes (averaged over 3 runs), using the task-agnostic evaluation protocol.}
\renewcommand\arraystretch{1}
\setlength{\tabcolsep}{1.97mm}
\scalebox{0.71}{
\begin{tabular}{@{}c|ccccccc|ccccccc@{}}
\toprule
\multirow{3}{*}{Method} & \multicolumn{7}{c|}{Accuracy of known classes (\%)}        & \multicolumn{7}{c}{Clustering accuracy of novel classes (\%)}          \\ \cmidrule(l){2-15} 
                        & \multicolumn{7}{c|}{Number of unlabeled classes}        & \multicolumn{7}{c}{Number of unlabeled classes}          \\
                        & 20   & 30   & 40   & 50   & 60   & 70   & 80   & 20   & 30   & 40   & 50    & 60   & 70   & 80   \\ \midrule
UNO \cite{fini2021unified}                     & 73.2 & 72.5 & 71.6 & 71.5 & 70.7 & 67.5 & 62.5 & 73.1 & 65.6 & 60.4 & 50.7  & 48.5 & 44.5 & 45.2 \\
IIC   \cite{li2023modeling}                  & 75.9 & 75.5 & 75.4 & 75.1 & 73.7 & 69.5 & 64.7 & 78.4 & 69.5 & 64.7 & 61.0  & 57.4 & 54.2 & 51.2 \\
rKD   \cite{peiyan2023class}                  & 75.2 & 75.2 & 75.6 & \textbf{78.6} & 74.2 & 71.4 & 69.6 & 76.4 & 65.7 & 61.7 & 59.4  & 55.6 & 52.6 & 47.5 \\
\textbf{SCKD}                    & \textbf{76.1} & \textbf{76.5} & \textbf{75.7} & 77.9 & \textbf{75.8} & \textbf{74.6} & \textbf{74.7} & \textbf{79.7} & \textbf{72.4} & \textbf{67.2} & \textbf{64.6} & \textbf{62.1} & \textbf{60.2} & \textbf{57.4} \\ \bottomrule
\end{tabular}
}
\label{tab4}
\end{table*}

\noindent \textbf{Varying the Number of Clusters.}
To compare the performance of different NCD methods in dealing with the imbalanced number of samples from known and novel classes, we set up a set of experiments with different proportions of the number of known and novel classes to approximate the situation of information imbalance between known and novel classes on CIFAR100.
Specifically, we divide the training set into seven ways, from 80/20 known/novel classes to 20/80 known/novel classes, and collect the prediction performance of various methods on the testing set for known and novel classes.
The results are shown in \cref{tab4}.
We observe that as the number of unlabeled classes gradually increases (20 $\to$ 80), 
existing methods gradually ignore the information of known classes, especially when they only account for a small proportion. 
An interesting phenomenon is that the model's performance for the novel classes also drops significantly even if the number of samples of the novel classes increases. 
We think a crucial reason is that the learning of novel classes relies on the knowledge of known classes, so the forgetting of known class knowledge also brings difficulties to the transfer of knowledge from known classes to novel ones.
In contrast, the performance of our method for known classes does not drop significantly.
Our method uses the information of all samples to learn known and novel classes to perform cooperative learning.
As a result, the stable and reliable performance proves that our SCKD method can better deal with the imbalance issue.

\begin{wraptable}{r}{6.0cm}
\centering
\vspace{-0.5cm}
\caption{Experiments with the estimated number of novel classes on CIFAR100-20, using the task-aware evaluation protocol.
}
\renewcommand\arraystretch{1}
\setlength{\tabcolsep}{3.5mm}
\scalebox{0.75}{
\begin{tabular}{@{}ccc@{}}
\toprule
\multirow{2}{*}{Method} & \multicolumn{2}{c}{Class number} \\ \cmidrule(l){2-3} 
                        & $C^u$=20              & $\hat{C^u}$=23             \\ \midrule
DTC \cite{han2019learning}                   & 67.3$\pm$1.2        & 64.3           \\
RankStats+  \cite{han2020automatically}            & 75.2$\pm$4.2        & 71.2           \\
UNO   \cite{fini2021unified}               & 90.5$\pm$0.7        & 71.5$\pm$1.8       \\
IIC \cite{li2023modeling}                    & 92.4$\pm$0.2        & 85.1$\pm$0.9       \\
\textbf{SCKD}                    & \textbf{92.6$\pm$0.6}        & \textbf{87.2$\pm$0.4}       \\ \bottomrule
\end{tabular}
}
\label{tab5}
\vspace{-0.3cm}
\end{wraptable}

\noindent \textbf{Unknown Number of Novel Classes.}
In the NCD task, the number of latent unlabelled classes $C^u$ is assumed to be known as a priori.
However, for real application scenarios, the number of classes (clusters) about the novel classes may be unknown.
Due to an insufficiently accurate number of novel classes, the model may over- or under-segment unknown unlabeled samples.
Considering this need, previous NCD methods (\eg, DualRank \cite{zhao2021novel}, UNO \cite{fini2021unified}, and IIC \cite{li2023modeling}) introduce an estimation algorithm proposed in DTC \cite{han2019learning} to estimate the number of unlabelled classes before the discovery stage.
Specifically, DTC first extracts a subset from the label set.
This subset is then clustered with the unlabelled set. 
After optimizing the clustering quality, the estimated number of novel classes can be obtained.
Following this setting, we test the performance of different methods on the estimated number of classes $\hat{C^u}$ on the CIFAR100-20 dataset split (averaged over 3 runs).
The results are shown in \cref{tab5}.
Compared with the accurate number of novel classes, our method achieves more significant gains (from 0.2 to 2.1) and stabler performance (smaller standard deviation) compared to other methods when dealing with the challenge of unknown number.
Our synthesized soft pseudo labels fuse the semantic information of multiple samples without depending on the number of novel classes, reducing adverse interference.
The results demonstrate that our method also performs best when facing more realistic challenges.

\subsection{Ablation Study}

We perform ablation study experiments on four dataset splits, including CIFAR100-20, CIFAR100-50, Stanford Cars, CUB, and FGVC-Aircraft.
In addition to the clustering accuracy introduced previously, we also introduce two commonly used clustering evaluation metrics: normalized mutual information (NMI) \cite{vinh2009information} and adjusted rand index (ARI) \cite{hubert1985comparing}, following existing methods \cite{fini2021unified,li2023modeling}.
MNI and ARI can measure the similarity between the clustering results and the ground-truth distributions (larger is better).

\noindent \textbf{Training Objectives $\mathcal{L}$.}
We also verify the proposed knowledge transfer objectives.
The results in \cref{tab6}(1)-(4) show the following observations.
First, using one of the proposed self-knowledge distillation objectives alone can also significantly improve the performance of the baseline. 
Combining all two objectives, the model can establish correlations between labeled and unlabeled samples, thereby achieving better results.
In addition, an interesting result is that compared to using known class information to directly guide novel class discovery $\mathcal{L}_{k\to n}$, using only novel class information to guide the review of known classes $\mathcal{L}_{n\to k}$ can significantly improve the model's performance in discovering novel classes on CIFAR100.
One reason may be that known and novel heads share the same trainable encoder. 
The process of using novel samples for learning the known classes also pays attention to the information of the novel classes and improves the encoder's ability to encode features of novel class samples.
For the Stanford Cars and CUB dataset, we freeze most encoder parameters.
In this case, using known information to guide novel classes can improve the model's performance in discovering novel classes more effectively.

\begin{table*}[t]
\centering
\caption{Ablation study of our method for four dataset splits on the unlabeled training subset, using task-aware evaluation protocol. Results contain clustering accuracy (ACC), Normalized Mutual Information (NMI), and Adjusted Rand Index (ARI) that are averaged over 3 runs. }
\renewcommand\arraystretch{1}
\setlength{\tabcolsep}{0.85mm}
\scalebox{0.675}{
\begin{tabular}{@{}c|ccc|ccc|ccc|ccc|ccc@{}}
\toprule
\multirow{2}{*}{ID} &
  \multirow{2}{*}{$\mathcal{L}_{k\to n}$} &
  \multirow{2}{*}{$\mathcal{L}_{n\to k}$} &
  \multirow{2}{*}{$E^r$} &
  \multicolumn{3}{c|}{CIFAR100-20} &
  \multicolumn{3}{c|}{CIFAR100-50} &
  \multicolumn{3}{c|}{Stanford Cars} &
  \multicolumn{3}{c}{CUB} \\
 &
   &
   &
   &
  ACC &
  NMI &
  ARI &
  ACC &
  NMI &
  ARI &
  ACC &
  NMI &
  ARI &
  ACC &
  NMI &
  ARI \\ \midrule
(1) &
  \XSolidBrush &
  \XSolidBrush &
   - &
  87.57 &
  0.8361 &
  0.7771 &
  62.62 &
  0.6861 &
  0.4742 &
  48.59 &
  0.6913 &
  0.3480 &
  63.56 &
  0.7869 &
  0.5002 \\
(2) &
  \Checkmark &
  \XSolidBrush &
  \Checkmark &
  91.57 &
  0.8617 &
  0.8333 &
  65.59 &
  0.6914 &
  0.4985 &
  54.19 &
  0.7270 &
  0.4029 &
  70.50 &
  0.8177 &
  0.5719 \\
(3) &
  \XSolidBrush &
  \Checkmark &
  \Checkmark &
    92.18 &
  0.8695 &
  0.8453 &
  65.98 &
  0.7069 &
  0.5165 &
  49.58 &
  0.6970 &
  0.3597 &
  68.77 &
  0.8040 &
  0.5486 \\
(4) &
  \Checkmark &
  \Checkmark &
  \Checkmark &
  \textbf{92.56} &
  \textbf{0.8754} &
  \textbf{0.8495} &
  \textbf{68.18} &
  \textbf{0.7128} &
  \textbf{0.5415} &
  \textbf{56.84} &
  \textbf{0.7388} &
  \textbf{0.4278} &
  \textbf{73.14} &
  \textbf{0.8262} &
  \textbf{0.5976} \\
(5) &
  \Checkmark &
  \Checkmark &
  \XSolidBrush &
  92.24 &
  0.8713 &
  0.8464 &
  66.35 &
  0.7094 &
  0.5256 &
  53.57 &
  0.7168 &
  0.3952 &
  69.78 &
  0.8074 &
  0.5635 \\ \bottomrule
\end{tabular}
}
\label{tab6}
\end{table*}

\noindent \textbf{Replica Encoder $E^r$.}
Since the known and novel classes share the same encoder.
Before discovering the novel stage, we deepcopy the pre-trained encoder to ensure unbiased feature information of the known classes.
We test the impact of the replica encoder $E^r $ on the results. When $E^r $ is not used, we use the target encoder itself to obtain the similarity scores between labeled and unlabeled samples.
The results in \cref{tab6}(4)-(5) show the following observations.
First, only using the target encoder itself can distinguish and associate labeled and unlabeled samples, making the proposed training target effective compared to the baseline.
In addition, the shared target encoder integrates information from all categories and is more likely to suffer from the imbalance of information between known and novel categories, especially when there are significantly more novel categories (\eg, CIFAR100-50).
The replica encoder retains pure prior information about known categories, thus bringing about a clearer similarity score calculation and promoting the model's differentiation of novel categories.

\noindent \textbf{Similarity Score Matrix $S$.}
The similarity score matrix $S$ is introduced to obtain the target pseudo-label by weighting the similarity of known and novel class features.
As shown in \cref{tab7}, we evaluate the effectiveness of our proposed 
\begin{wraptable}{r}{6.0cm}
\centering
\caption{Ablation study about the similarity score matrix $S$. All results are evaluated on the unlabeled training set.}
\setlength{\tabcolsep}{1.5mm}
\scalebox{0.75}{
\begin{tabular}{@{}c|ccc@{}}
\toprule
Settings  & Stanford Cars & CUB  &  FGVC-Aircraft \\ \midrule
Average $S$ & 49.3          & 64.7   &      52.6     \\
Random $S$  & 47.9          & 62.8   &      50.2     \\
\textbf{Our}       & \textbf{56.8}          & \textbf{73.1} & \textbf{56.5}     \\ \bottomrule
\end{tabular}
}
\label{tab7}
\vspace{-0.3cm}
\end{wraptable}
similarity weighting method.
`Average $S$' denotes all samples have the same influence coefficients.
`Random $S$' denotes all samples randomly generate influence coefficients ranging from 0 to 1.
Our weighting method significantly improves model performance.
`Average $S$' causes the model to be unable to focus on the mutual information between categories, causing the model to be unable to fully utilize the sample information.
`Random $S$' may produce pseudo correlations among samples and interfere with model learning.

\begin{wrapfigure}{r}{7cm}
	\centering
    \vspace{-0.5cm}
	\includegraphics[scale=0.3]{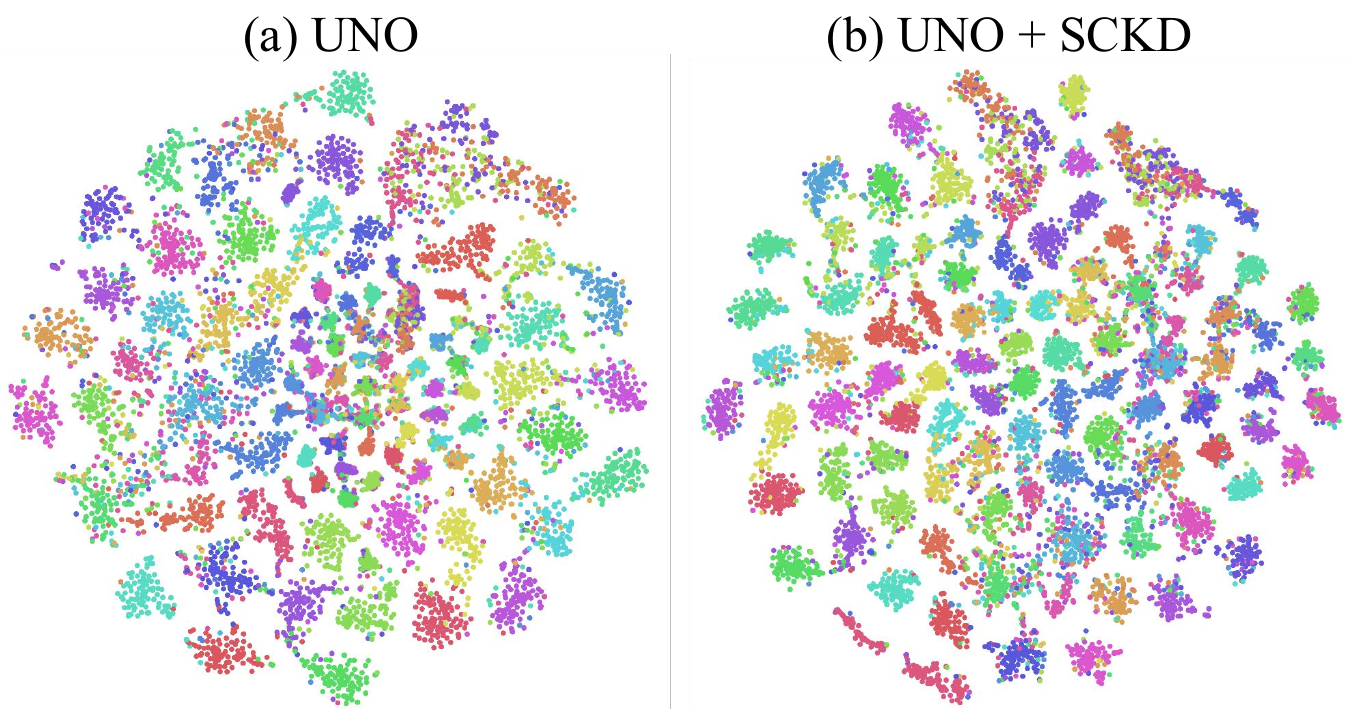}
	\caption{t-SNE visualization for known and novel classes on CIFAR100-50 testing set, using task-agnostic evaluation protocol.}
	\label{fig4}
	\vspace{-0.5cm}
\end{wrapfigure}

\subsection{Visualization}
To intuitively verify the effectiveness of the method, we further provide qualitative results.
As shown in \cref{fig4}, we use the t-SNE tool \cite{van2008visualizing} to observe the model's prediction on the CIFAR100-50 testing set.
Compared with the original UNO \cite{fini2021unified}, combining with the proposed SCKD can more tightly group samples of the same classes.
In accordance with the visualization, SCKD helps to separate the classes.

\section{Conclusion}
This paper considers a practical but long-neglected challenge in the Novel Class Discovery (NCD) task, \ie, the imbalanced number of samples from known and novel classes.
The model despises the learning of novel classes when the known class dominates while the model forgets the prior knowledge of the known class when the novel class dominates.
Compared to transferring knowledge by a shared representation space, we collect the model’s feature representations of known and novel classes as two disjoint representation spaces.
Then we propose a Self-Cooperation Knowledge Distillation (SCKD) method to utilize every sample for review and discovery.
Our method is conceptually simple and intuitive and achieves top-level competitive results in numerous experiments on six datasets.

\vspace{0.5cm}

\noindent\textbf{Acknowledgements.} This work is supported by the Engineering Research Center of AI \& Robotics, Ministry of Education, China.


%
%
\bibliographystyle{splncs04}
\bibliography{main}

\newpage

\begin{center}
\linespread{1.6}\selectfont
{\Large{\textbf{\textit{Supplementary Material for} Self-Cooperation Knowledge Distillation for Novel Class Discovery}}}
\end{center}

\vspace{0.8cm}

In this supplementary material, we provide more details
of our method, organized as follows:

\begin{itemize}
    \item In Section~\ref{sec1}, we further discuss related but different tasks of NCD, mainly in terms of training goals and training \& testing processes, including OSR and GCD tasks.
    \item In Section~\ref{sec3}, we show the hyperparameter experiment, corresponding to Section \textcolor{blue}{4.1} of the main body.
    \item In Section~\ref{sec2}, we compare the performance of various NCD methods when facing the cross-domain dataset to measure the effectiveness of information transfer between known and novel classes.
    \item In Section~\ref{sec4}, we reassess the model's performance on three fine-grained datasets without the pre-trained model.
    \item In Section~\ref{sec6}, we evaluate the trade-offs between the two parts of the proposed Self-Cooperation Knowledge Distillation losses and provide readers with a reference for setting parameters according to actual needs.
    \item In Section~\ref{sec7}, we discuss the differences between our and other NCD methods.
    \item In Section~\ref{sec8}, we provide insights about future works.
    \item In Section~\ref{sec9}, we discuss the limitations of the proposed method.
\end{itemize}

\setcounter{section}{0}

\section{Discussion of Related but Distinct Tasks}
\label{sec1}

\textbf{Open-Set Recognition (OSR)} \cite{6365193} aims to detect test-time images which do not belong to one of the classes in the labeled set, but does not require any further classification amongst these detected images.
Therefore, the model only needs to maintain a high degree of confidence in the known class, and does not consider the model's scalability for the novel class.

\noindent \textbf{Novel Class Discovery (NCD)} \cite{hsu2017learning,hsu2019multi} not only considers the review of known classes, but also considers the discovery of unknown classes to expand the scope of application of the model.
More detailed settings and definitions are shown in the main body.

\noindent \textbf{Generalized Category Discovery (GCD)} \cite{vaze2022generalized} is similar to NCD.
In the training process, GCD and NCD contains the same dependencies, \ie, a labeled set of known classes and an unlabeled set of unknown classes.
The difference is that GCD makes no assumptions about the test set. Test samples can come from the known or novel classes.

\begin{table}[t]
\centering
\caption{Analysis of hyperparameter $\beta$ and $\alpha$. ``Train-Novel" refers to the evaluation of the unlabeled training subset, using task-aware evaluation protocol. 
Other results are evaluations of the testing subset, using the task-agnostic evaluation protocol.}
\renewcommand\arraystretch{1}
\setlength{\tabcolsep}{2.5mm}
\scalebox{0.95}{
\begin{tabular}{@{}c|cc|cccc@{}}
\toprule
\multirow{2}{*}{ID} & \multicolumn{2}{c|}{\multirow{2}{*}{Setting}} & \multicolumn{4}{c}{CIFAR100-50}                  \\
                    & \multicolumn{2}{c|}{}                         & Known & Novel & All & Train-Novel \\ \midrule
(1)  & \multirow{5}{*}{$\beta$ }  & 0    & 70.3 & 49.7 & 60.0   & 62.6 \\
(2)  &                        & 0.01 & 70.7 & 50.6 & 60.7 & 63.3 \\
(3)  &                        & 0.1  & \textbf{78.1} & 56.4 & 67.3 & 64.7 \\
(4)  &                        & 0.5  & 77.9 & \textbf{62.6} & \textbf{70.3} & \textbf{68.2} \\
(5)  &                        & 1    & 77.3 & 61.2 & 69.3 & 66.3 \\ \midrule
(6)  & \multirow{5}{*}{$\alpha$} & 0.01 & 74.5 & 51.8 & 63.2 & 63.6 \\
(7)  &                        & 0.05 & 77.5 & 61.2 & 69.4 & 67.2 \\
(8)  &                        & 0.1  & \textbf{77.9} & \textbf{62.6} & \textbf{70.3} & \textbf{68.2} \\
(9)  &                        & 0.5  & 75.4 & 59.4 & 67.4 & 65.8 \\
(10) &                        & 1    & 76.4 & 58.5 & 67.5 & 65.4 \\ \bottomrule
\end{tabular}
}
\label{tab2_sup}
\end{table}

\section{Hyper-parameter Experiment}
\label{sec3}

In this section, we show the test results of the hyperparameter $\beta$ and $\alpha$.
The results are shown in Table~\ref{tab2_sup}.
We test on the CIFAR100-50 dataset split and report the results on the training and testing subsets.
For (1)-(5), we fix the $\alpha$ as 0.1 and show the results with various $\beta$.
Compared with the original version $ \beta=0$, the proposed SCKD technology can help the model improve the performance of known and novel classes.
When the value of $\beta$ is too high, the model may be overly encouraged to focus on another representation space and underestimate its original pure knowledge (the labeled samples for known classes \& unlabeled samples for novel classes).
We recommend choosing $ \beta=0.5$ as the default value for better results.
For (6)-(10), we fix the $\beta$ as 0.5.
Different $\alpha$ affects pseudo-labels' smoothness, thereby achieving different degrees of mutual information interference between known and novel classes.
By default, we choose $\alpha=0.1$.

\section{Experiments on Cross-domain Dataset}
\label{sec2}

In the main body of the paper, we have evaluated the performance of the algorithm on disjoint known and novel classes.
In order to verify the effectiveness of the method more broadly, we conduct experiments when the known and novel classes satisfy the cross-domain yet category related.
We select Office-31 \cite{saenko2010adapting}, a well-known domain adaptation evaluation dataset with 31 categories of office objects.
It contains 4652 images from three domains: Amazon (A), DSLR (D), and Webcam (W).
We experiment with all six combinations (source domain $\to$ target domain): A$\to$W, D$\to$W, W$\to$D, A$\to$D, D$\to$A, W$\to$A, and report the average accuracy based for each baseline.
Following KCL \cite{hsu2017learning}, we use ResNet-18 as the backbone pre-trained with ImageNet.
Each mini-batch is constructed by 32 labeled samples from source domain and 96 unlabeled samples from target domain.
The results are shown in Table~\ref{tab1_sup}, our method achieves state-of-the-art results and outperforms the previous best baseline method by an average accuracy of 2.5\% to 9.8\%.
By representing the mutual information between spaces and the cooperation learning paradigm, our method can make more full use of limited sample information, thereby achieving competitive results.

\begin{table}[t]
\centering
\caption{Cross-domain NCD on the Office-31 dataset. Avg represents the average accuracy.}
\renewcommand\arraystretch{1}
\setlength{\tabcolsep}{2.5mm}
\scalebox{0.95}{
\begin{tabular}{@{}c|ccccccc@{}}
\toprule
Method & A$\to$W  & D$\to$W  & W$\to$D  & A$\to$D  & D$\to$A  & W$\to$A  & Avg  \\ \midrule
KCL\cite{hsu2017learning}    & 76.7 & 97.3 & 98.2 & 71.2 & 61.0 & 60.5 & 77.5 \\
UNO\cite{fini2021unified}    & 85.6 & 97.3 & 98.6 & 74.6 & 66.4 & 65.7 & 81.4 \\
IIC\cite{li2023modeling}    & 89.4 & 98.7 & 99.0 & 79.8 & 70.6 & 71.4 & 84.8 \\
\textbf{SCKD}   & \textbf{90.4} & \textbf{99.1} & \textbf{99.8} & \textbf{84.6} & \textbf{74.4} & \textbf{75.4} & \textbf{87.3} \\ \bottomrule
\end{tabular}
}
\label{tab1_sup}
\end{table}

\begin{table}[h]
\centering
\caption{Experiments on fine-grained datasets without the DINO pre-trained ViT model.}
\renewcommand\arraystretch{1}
\setlength{\tabcolsep}{3mm}
\scalebox{0.95}{
\begin{tabular}{@{}cccc@{}}
\toprule
Methods     & Stanford Cars & CUB  & FGVC-Aircraft \\ \midrule
RankStats+ \cite{han2020automatically} & 17.7          & 20.2 & 30.1          \\
NCL \cite{zhong2021neighborhood}       & 27.1          & 24.7 & 36.7          \\
UNO \cite{fini2021unified}       & 25.2          & 24.5 & 37.8          \\
rKD \cite{peiyan2023class}      & 29.2          & 26.5 & 43.8          \\
\textbf{SCKD}       & \textbf{31.2}          & \textbf{29.7} & \textbf{48.5}          \\ \bottomrule
\end{tabular}
}
\label{tab3_sup}
\end{table}

\section{Experiments without Pre-trained Model}
\label{sec4}

For three fine-grained datasets (\ie, Stanford Cars, CUB, and FGVC-Aircraft), we choose the pre-trained ViT-B-16 model, following GCD \cite{vaze2022generalized} and rKD \cite{peiyan2023class}.
In this section, we utilize ResNet-18 as the backbone from scratch.
We first pre-train the model on known classes for 200 epochs and then train on known and novel classes for 200.
The results are shown in Tabel~\ref{tab3_sup}
For the three datasets, our method improves by 2.0\%, 3.2\%, and 4.7\%, respectively.
The results demonstrate our method's effectiveness without the pre-trained models.

\section{Trade-off between Self-Cooperation Knowledge Distillation Losses}
\label{sec6}

As \textcolor{blue}{Eq.(7)} of the main body, our overall Self-Cooperation Knowledge Distillation loss consists of two parts. 
The information on labeled samples assists in the discovery of novel classes, and the information of unlabeled samples assists in the review of known classes as:
$\mathcal{L}_{SCKD} = \mathcal{L}_{k\to n} + \mathcal{L}_{n\to k}.$
\ie, $\mathcal{L}_{SCKD} = 2 \times [\lambda \cdot \mathcal{L}_{k\to n} + (1-\lambda )\cdot \mathcal{L}_{n\to k}], \lambda =0.5.$
In this section, we use $\lambda$ as an adjustable parameter to balance the two.
The results are shown in Table~\ref{tab5_sup}.
We find that adjusting $\lambda$ can affect the review of known classes and the discovery of novel classes to a certain extent. Specifically, when $\lambda$ is small, the model is encouraged to review known classes from unlabeled sample information, thus obtaining higher performance on known classes.
When $\lambda$ increases, the model pays more attention to using labeled sample information to discover novel classes. 
Overall, adjusting $\lambda$ has little impact on the overall accuracy. 
For convenience, we set $\lambda=0.5$ and encourage changing the size of $\lambda$ as needed.

\begin{table}[t]
\centering
\caption{Experimental results with different $\lambda$ on CIFAR100-50 testing subset, using task-agnostic evaluation protocol.}
\renewcommand\arraystretch{1}
\setlength{\tabcolsep}{3mm}
\scalebox{1}{
\begin{tabular}{@{}cccccc@{}}
\toprule
$\lambda$ & $\mathcal{L}_{k\to n}$   & $\mathcal{L}_{n\to k}$   & Known & Novel & All  \\ \midrule
0.1                    & 0.1 & 0.9 & 78.4  & 60.6  & 69.5 \\
0.3                    & 0.3 & 0.7 & 78.0  & 61.2  & 69.6 \\
0.5                    & 0.5 & 0.5 & 77.9  & 62.6  & 70.3 \\
0.7                    & 0.7 & 0.3 & 76.9  & 62.4  & 69.7 \\
0.9                    & 0.9 & 0.1 & 74.6  & 64.6  & 69.6 \\ \bottomrule
\end{tabular}
}
\label{tab5_sup}
\end{table}

\section{Discussion with NCDLR \cite{zhang2023novel} and rKD \cite{peiyan2023class}}
\label{sec7}

In NCDLR \cite{zhang2023novel}, the author considers the long-tail distribution problem of data between novel classes and introduces an equiangular prototype representation method to provide pseudo labels for novel classes.
The problem exists among novel classes without considering the distribution of the number of all training samples in the population.
Unlike their motivation, our work considers another crucial real-life issue in NCD tasks.
We focus on the balance of information between known and novel classes.
In fact, deep learning models are often expected to continue to expand their scope of application. 
In this process, the problem of the imbalanced number of samples from known and novel classes is inevitable. 
In rKD \cite{peiyan2023class}, the author encourages the model to maintain the same proportional relationship between the predictions of known classes and the discovery stage of novel classes.
In this process, their method weakens the information transfer process between known and novel classes. 
At the same time, the method does not consider the distribution differences of different categories of data. 
When the distribution shifts, the performance will be significantly affected (refer to Table~4 of the main body).
Our method is motivated by trying to compensate for a missing perspective in current tasks, namely the imbalanced number of samples from known and novel classes.
Besides, our proposed method collaboratively reviews known classes and discovers novel classes by combining information from all labeled and unlabeled samples. 
The motivation and method are significantly different from existing NCD baselines.

\section{Future Works}
\label{sec8}

In this work, we consider a real application challenge in NCD tasks, \ie, the imbalanced number of samples from known and novel classes.
In fact, the scalable foundation model \cite{kirillov2023segment} also faces similar needs and challenges.
When the foundation model is required to expand new application scenarios, learning about unknown objects is often accompanied by catastrophic forgetting of known knowledge.
How to fully utilize the information of all samples to enhance the performance of such large foundation models with special needs remains to be solved.
We leave this section as future work.

\section{Limitations}
\label{sec9}

Our method is able to cope with the challenge of the imbalanced number of samples from known and novel classes and achieves considerable progress.
But it is still not perfect.
An obvious limitation is that our method assumes that known and novel samples can be few but not completely absent. 
Like other NCD methods based on information transfer, in this extreme case, our method and other methods degenerate into a simple coupling of supervised learning and unsupervised clustering.
For such extreme cases, additional information banks may be needed for information storage before deploying our proposed technology.

\end{document}